\pdfoutput=1

\documentclass[11pt]{article}

\usepackage[final]{coling}

\usepackage{times}
\usepackage{latexsym}

\usepackage[T1]{fontenc}

\usepackage[utf8]{inputenc}

\usepackage{microtype}

\usepackage{inconsolata}

\usepackage{graphicx}

\usepackage{amsmath}
\usepackage{amsthm}
\usepackage{graphicx}
\usepackage{comment}
\usepackage{multirow}
\usepackage{enumitem}
\usepackage{booktabs}
\usepackage{tikz}
\usepackage{linguex}
\usepackage{array}
\usepackage{listings}
\usepackage{float}
\DeclareMathOperator*{\argmin}{\arg\!\min}
\theoremstyle{definition}
\newtheorem{definition}{Definition}[section]
\newcolumntype{S}{>{\small}c}

\newcommand{\emily}[1]{{\color{orange}[#1]$_{EA}$}}

\title{Generics are puzzling.\\ Can language models find the missing piece?}

\author{Gustavo Cilleruelo Calderón \phantom{aaa} Emily Allaway \phantom{aaa} Barry Haddow \phantom{aaa} Alexandra Birch \\
School of Informatics, University of Edinburgh\\
\texttt{\small g.cilleruelo-calderon@sms.ed.ac.uk \phantom{aaa} \{emily.allaway, bhaddow, a.birch\}@ed.ac.uk}}

\begin{document}
\maketitle
\begin{abstract}
Generic sentences express generalisations about the world without explicit quantification. Although generics are central to everyday communication, 
building a precise semantic framework has proven difficult, in part because speakers use 
generics
to generalise properties with widely different statistical prevalence. 
In this work, we study the implicit quantification and context-sensitivity of generics by leveraging language models as models of language. 
We create \textsc{ConGen}, a dataset of 2873 naturally occurring generic and quantified sentences in context, and define \textit{p-acceptability}, a metric based on surprisal that is sensitive to quantification.
Our experiments show generics are more context-sensitive than determiner quantifiers and about 20\% of naturally occurring generics we analyze express weak generalisations. We also explore how human biases in stereotypes can be observed in language models\footnote{Code and data are available in \href{https://github.com/ilyocoris/generics_are_puzzling}{\texttt{https://github.com/ilyocoris/generics\_are\_puzzling}}}.
\end{abstract}

\section{Introduction}

Humans use generalisations to
abstract away from particular objects, events or facts and convey regularities about the world.

In this work, our focus is on generic sentences, such as \textit{insects have six legs} or \textit{mosquitoes carry malaria}, which express generalisations without explicit quantification. These two generic sentences are acceptable in many contexts, but the quantifications they 
convey are
widely different: almost all insects have six legs, but fewer than 1\% of mosquitoes carry malaria.

One way of expressing generalisations in language is through explicitly quantified sentences, such as 
\textit{most insects are nocturnal} or \textit{some mosquitoes have white stripes}. 
Quantified sentences express statistical claims about the members of a kind that share the predicated property: for example \textit{most} if a majority of insects are nocturnal or \textit{some} if a minority of mosquitoes have white stripes.


Even as generics seem to express 
inconsistent
quantifications, they are at the heart of communication and
dissemination
in science~\citep{dejesus2019gen_science,bowker2022gen_scientific_communication},
medical research~\citep{peters2024medicalgenerics}, and politics~\citep{novoa2023generics_polarization}. Furthermore, in the social realm generics
serve as linguistic vehicle for
social essentialism  \citep{rhodes2012cultural} and stereotyping~\citep{leslie2017originalsin,bosse2022stereotyping}.

The 
nature of generics and their importance in communication has 
led to
extensive literature 
on
the semantics of generic sentences~\citep[e.g.,][]{carlson1977disseration,cohen1999probability,generics_08_leslie,liebesman2011simple,sterken2015context,nickel2016between,goodman2018generalization,stovall2019tickets,nguyen2020radical,bosse2021nonspecific,kirkpatrick2023dynamics}. However, many open questions remain.
These include how they relate to quantifiers and the degree to which generics are context sensitive.
In this work, we use language models to explore the implicit quantification and context-sensitivity of generics, and how they are affected by human biases around stereotypes. 


Language models have demonstrated unprecedented performance in a variety of linguistic tasks, such as machine translation \citep{kocmi2024wmt} or conversational assistance \citep{chiang2024chatbotarenaopenplatform}. 
We describe how speakers use generics by studying the surprisal in language models for various naturally occurring generic and quantified sentences.


Most existing datasets of generics 
are synthetic,
often derived from knowledge bases or generated by language models~\citep{Bhakthavatsalam_2020_genericskb, allaway2024exemplars_llms}. 
Since the examples in these datasets are machine-generated and/or lack a context in which they might be uttered,
there is no guarantee that they represent how
speakers
actually use generics.
Therefore, in this work we introduce \textsc{ConGen}, a dataset of naturally occurring 
generic and quantified sentences with contexts.

In order to 
study
generics and quantification, we 
define the p-acceptability metric;
given a set of quantifiers, it indentifies the one that best fits a sentence by using the surprisal of a language model. Previous work either fails to use surprisal to describe quantification or focuses on prompting  \citep{collacciani2024quantifying,allaway2024exemplars_llms}.
We 
validate this metric by showing that it
recovers the expected dynamics of 
quantifiers (\textit{all}, \textit{most}, \textit{some}) and the generic on two datasets of generics (\textsc{ConGen} and \textsc{GenericsKB}). We then use our metric to study different aspects of generics.

Our contributions are (\emph{i}) \textsc{ConGen}, a dataset of naturally occurring bare plural generic and quantified sentences in contexts, (\emph{ii}) p-acceptability, a new metric based on language models that is sensitive to quantification and (\emph{iii}) insights into how generics are used, including weak generalisations, context-sensitivity and stereotypes.

\section{Background}
\label{sec:background}

Semantic theories of generics guide and scaffold our experimental design. 
In what follows, we present 
linguistic background on generics (\S\ref{sec:generics})
and then
we sketch two theories of genericity related to our experiments (\S\ref{sec:semantic_theories}): \textit{generics-as-defaults} and \textit{contextualism}. 
Finally, we introduce two 
phenomena that involve generics (\S\ref{sec:generics_and_human_bias}):
\textit{stereotypes} and \textit{generic overgeneralisation}.
These theoretical elements motivate our research questions and frame the interpretation of experimental results.

\subsection{Generics}
\label{sec:generics}
The term
\textit{generics} covers multiple 
linguistic phenomena 
that
abstract away from particular objects, members, or events.
In our work, we focus on one specific kind of generics: \textit{bare plural characteristic sentences}. Bare plurals are noun phrases in plural form without a definite or indefinite article\footnote{\textit{Sharks attack bathers} is a bare plural generic. The same generic could also be expressed in English with the definite (\textit{the shark attacks bathers}) or indefinite (\textit{a shark attacks bathers}) articles.}. Characteristic sentences are propositions that do not express specific episodes or isolated facts, but rather report a kind of general property or regularity \citep{carlson1995thegenericbook}. 

In linguistics, generics are traditionally analyzed as quantifiers \citep{carlson1977disseration}, with an unpronounced implicit operator \textsc{Gen} that 
has
a role similar to that of \textit{most} or \textit{generally} in explicit quantification \citep{lewis1975adverbs}. However, there is no consensus on what the semantic content of \textsc{Gen} is, how it is determined or even if it exists~\citep[cf.][]{carlson1995thegenericbook}. 
Despite this,
most real-life generics are \textit{majority generics}: they are acceptable when a majority of the members of the kind in question satisfies the predicated property (e.g. \textit{ravens are black}).

However, some generics, such as \textit{mosquitoes carry malaria}, \textit{ducks lay eggs}, or \textit{bees reproduce}, are often used
even though
they express statistically weak generalisations. We call these \textit{weak generics}~\citep{weak_22_almotahari}. 
Weak generics are broadly categorized into two types \citep{leslie2007structure}: those that 
express properties that are characteristic of the kind, but are only possessed by a minority of its members (\textit{minority generics}) and those that express dangerous, striking or appalling characteristics (\textit{striking generics}).

In addition to the above, \citet{leslie2011gog} also distinguish
\textit{quasi-definitional} generics. These predicate a property true of all the members of the kind, without exceptions (e.g. \textit{beetles are insects}). 

\subsection{Generics in Philosophy of language}
\label{sec:semantic_theories}

In this section we introduce two influential accounts from philosophy of language: generics-as-defaults and contextualism. These supply contrasting views on how to explain the diversity in generic use and how they are affected by context; topics we discuss in our experiments (\S\ref{sec:implicit_quantification} and \S\ref{sec:context_sensitivity} respectively).

\paragraph{Generics as defaults.}
The \textit{generics-as-defaults} theory posits generics as the linguistic manifestation of a default cognitive mechanism of generalisation~\citep{generics_08_leslie}. 
In contrast to quantifiers, which express generalisations based on statistical surveying, generics express primitive generalisations based on what we perceive as characteristic, distinctive or striking in the world \citep{leslie2007structure}. 

\paragraph{Contextualism.}
\citet{sterken2015context} argues for a contextualist view of generics: generics express widely different generalisations in different contexts. The unpronounced generic operator \textsc{Gen} picks out a generalisation as a function of the context of the utterance, similarly to how the determiner \textit{that} picks out a referent.




\subsection{Human biases in the usage of generics}
\label{sec:generics_and_human_bias}

Generics are often used in ways that do not follow logical reasoning and highlight human cognitive biases \citep{leslie2017originalsin,neufeld2025giving}. One important example is the connection generics have with how we express stereotypes; we explore this in experiment \S\ref{sec:striking}.


\paragraph{Generic overgeneralisation.}  One example of illogical use of generics is \textit{generic overgeneralisation} \citep{leslie2011gog,lazaridou2017gog}: humans often use universal quantification (\textit{all}) in situations where the generic is acceptable even when exceptions exist.
This effect has also been documented in language models~\citep{allaway2024exemplars_llms,ralethe-buys_2022_generic}. 

\paragraph{Stereotypes.} In the social realm, striking generics are linked to stereotyping and the essentialization of social groups~\citep{rhodes2012cultural,leslie2017originalsin}. 
In particular, 
\citet{cimpian2010generic} and \citet{leslie2009gen_experiment} demonstrate a psychological connection between striking information and an overestimation of statistical frequencies. 
This means that humans seem to reason from the quantifier
\textit{some} to \textit{most} and even to \textit{all} when striking properties are at play~\citep{generics_08_leslie}.
Additionally, generics are also 
central in
recent NLP studies on preventing and countering stereotypes~\citep{bosse2022stereotyping,allaway2023countering,mun2023stereotypes}. 

\section{Related work}
\label{sec:related_work_nlp}




Recent works that study generics and language models use prompting to test generic overgeneralisation, property inheritance~\citep{allaway2024exemplars_llms} and, more generally, the effect of quantifiers on sentence meaning \citep{collacciani2024quantifying}. However, for studying how language models model quantification and generics, prompting has several shortcomings. In particular, the effect of small variations in the prompt on model behavior is not well understood~\citep{salinas2024butterflyprompts}.
Additionally, prompting requires an instruction tuned model, often trained to be a virtual assistant~\citep{zhang2024instructiontuninglargelanguage}, which may skew the underlying language distribution in unaccounted ways. 

To avoid the drawbacks of prompting, studies have also looked at the internal states of pre-trained models. 
\citet{collacciani2024quantifying} compare the surprisals of quantified sentences but fail to find a 
sensitivity to quantification in language models. As the authors note, this may be due to their metric not being sufficiently expressive. 
While the work from \citet{gupta2023quantifier_comprehension} also uses surprisal, in this case of critical words,
to draw conclusions about quantifier comprehension in language models, it does not take generics into consideration.
In contrast, 
in this work we develop a new metric that uses the surprisal 
of the predicated property tokens, and show that it describes rich quantificational dynamics modelled by language models.


Several datasets exist that specifically target generics. 
\textsc{GenericsKB} is a dataset ($3$M samples) that combines naturally occurring generic and quantified statements
with synthetic examples derived from knowledge bases~\citep{Bhakthavatsalam_2020_genericskb}. The naturally occurring generics are selected with a \textsc{Bert}-based scorer trained on human annotations.
The \textsc{Gen-A-Tomic} corpus contains synthetic generics generated by \textsc{Gpt2-xl}~\citep{bhagavatula2023i2d2}.
Additionally, datasets of synthetic generics exemplars (i.e., cases where the generic does and does not hold) have been constructed~\citep{allaway2023penguins, allaway2024exemplars_llms}.

All of these datasets 
contain synthetic examples (either machine generated or derived from knowledge bases) and do not include context, which is key to understanding how speakers use generics.
In contrast, our \textsc{ConGen} dataset contains only naturally occurring human-annotated sentences, each with an associated document as context.


\section{Methodology}
\label{sec:methodology}

\subsection{Dataset: \textsc{ConGen}}
\label{sec:congen}

\begin{table*}
  \small
  \centering
  \begin{tabular}{p{7.5cm} p{7.5cm}}
   \multicolumn{1}{c}{\normalsize{\textsc{Dolma}}} & \multicolumn{1}{c}{\normalsize{\textsc{Reddit 2024}}} \\
  \toprule
  The potato variety dictates the color of the flower which for red potatoes can be dark pink to lavender. \textbf{Yellow potatoes have white flowers.}
Piling our storage potatoes starts in late-September and by mid-October all our potatoes are in the barn. & 
No. Sprinters typically have long legs. Runners in general have long legs. \textbf{Swimmers have long torsos.} Michael Phelps, who is 6'4", has shorter legs than the Olympic runner Hicham El Guerrouj, who is 5'9" \\
\midrule
There are several reasons for a high number of repetitive leg movements while sleeping. \textbf{Some people with chronic pain at night tend to have poor sleep and frequent repetitive leg movements.} If you have concerns about your sleep, you should discuss them with your doctor.
& I have this problem. It’s not because I don’t like vegetables. I can just taste waaaay too much of the minerals. \textbf{Most vegetables taste like iron and dirt.} I’d rather eat actual dirt than a beet. Also it makes water taste weird when you can taste minerals.\\
  
  \bottomrule

  \end{tabular}
  \caption{Examples of generic and quantified sentences in context, extracted from the \textsc{ConGen} dataset.}
  \label{tab:congen_examples}
\end{table*}

Theorists emphasize the context-sensitivity of generic sentences~\citep{sterken2015context,nickel2016between,gencontext_23_almotahari}. The lack of consensus on how context affects the use of generics motivates the construction of \textsc{ConGen}. To the best of our knowledge, this is the first dataset that targets generics in context. 

\textsc{ConGen} consists of naturally occurring bare plural generics and quantified statements (with \textit{some}, \textit{most} and \textit{all}) in context. 
These are drawn from a subset of \textsc{Dolma} \citep{soldaini2024dolma} and from 2024 Reddit comments.
\textsc{Dolma} is a cleaner version of Common Crawl and may have been used in the training data for popular language models (e.g., \textsc{Mistral}). Therefore, 
we include recent Reddit comments to validate our findings on data the models have not been trained on.

\begin{table}[!t]
  \small
  \centering
  \begin{tabular}{lll}
   Source & Quantifier & \# Samples \\
   \midrule
   \multirow{4}{*}{\textsc{Dolma}} & \small\textsc{Gen} & 559 \\
                      & \small\textsc{All} & 500 \\
                      & \small\textsc{Most} & 578 \\
                      & \small\textsc{Some} & 551 \\
   \midrule
   \multirow{4}{*}{Reddit (2024)} & \small\textsc{Gen} & 411 \\
                      & \small\textsc{All} & 71 \\
                      & \small\textsc{Most} & 158 \\
                      & \small\textsc{Some} & 45 \\
  \end{tabular}
  \caption{\textsc{ConGen} dataset: breakdown of the 2873 annotated sentences in context.}
  \label{tab:congen}
\end{table}

In order to find bare plural generic sentences in such massive collections of data, we train a binary classifier to detect generic and quantified sentences. We train a \textsc{RoBERTa} \citep{liu2019roberta} classifier on \textsc{GenericsKB} and \textsc{Gen-A-Tomic}. 

We find candidate sentences in the original data sources by combining the scores of the classifier with linguistic heuristics that filter out sentences in the singular or in past tense. The collected candidate sentences are annotated  as irrelevant, bare plural generic or explicitly quantified with \textit{all}, \textit{most} or \textit{some}. The final dataset contains $2873$ human-annotated generic and quantified sentences (Table \ref{tab:congen}).
Details on dataset construction are available in Appendix \ref{apx:congen}.

\subsection{Metric: p-acceptability}
\label{sec:acceptability}

Generic sentences can be used to express generalisations with vastly different quantificational strength: from weak generics (e.g., \textit{mosquitoes carry malaria}) to quasi-definitional ones (e.g., \textit{mosquitoes are insects}).
To describe and study these quantificational dynamics
in language models, we introduce a criterion 
to answer the following question:
What is the quantifier that best fits the kind-property relation expressed in a sentence?

Consider quantified bare plural generalisations with a simple structure (context $+$ quantifier $+$ bare plural $+$ verb $+$ property), where the quantifier is one of \textit{all}, \textit{most}, \textit{some} or the generic (\O). We propose a notion of acceptability that selects the \textit{quantifier that makes the property more likely} given the subject, verb and context.

\begin{definition}[p-acceptability]
  Let $Q$ be a set of candidate quantifiers, $s$ a bare plural generic and $\theta$ a language model. We construct 
  $\{q+s\hspace{3pt}|\hspace{3pt}q\in Q\}$ the set of variations of $s$\footnote{$q+s$ refers to string concatenation: if $q=\text{most}$ and $s=\textit{tigers have stripes}$, then $q+s=\text{most tigers have stripes}$.}. We call $q$ the p-acceptable quantifier for $s$ if $q+s$ is the sentence with the lowest surprisal of the property tokens (i.e. tokens after the verb):
  \begin{equation}
    \label{eq:p-acceptability}
    \text{p-acceptable}(s;Q, \theta) := \argmin_{q\in Q} H_p(q+s;\theta)
  \end{equation}
where $H_p$ is the surprisal of the property tokens
  \begin{equation}
    H_p(s;\theta) := -\frac{1}{|P|}\sum_{i\in P}\log p_\theta(t_i|t_{< i})
  \end{equation}
with $P$ is the set of indices of the property tokens and $t_i$ the tokens in sentence $s$. 
\end{definition}
We build the set of variations of $s$ as $\{s, $ `all'$+s$, `most'$+s$, `some'$+s$$\}$.
For sentences that originally had an explicit quantifier, 
we remove the quantifier to obtain $s$. To compute the p-acceptability of a sentence $s$ with context $c$, we build the set of variations $\{c+q+s\hspace{3pt}|\hspace{3pt}q \in Q\}$.

For example, consider the sentence $s=\textit{tigers have stripes}$ 
which can be split into word tokens 
$t_0=\texttt{tigers}$, $t_1=\texttt{have}$ $t_2=\texttt{stripes}$. Recall that the candidate quantifiers are $Q=\{\text{all},\text{most},\text{some},\text{\O}\}$. Then, the set of variations will be \textit{all tigers have stripes}, \textit{most tigers have stripes}, \textit{some tigers have stripes} and \textit{tigers have stripes}. The surprisal on the property tokens (in this case $t_2$) with the quantifier ``all'' is then calculated as:
\begin{multline*}
  H_p(\texttt{all}+\texttt{tigers have stripes};\theta) = \\
 -\frac{1}{|P|}\sum_{i\in P}\log p_\theta(\texttt{stripes}|\texttt{all tigers have})
\end{multline*} and similarly for the other quantifiers in $Q$. The p-acceptable quantifier would then be the one with the minimum surprisal.

As we show in the following experiments, p-acceptability is sensitive to the effect of using a quantifier (\textit{all}, \textit{most}, \textit{some}) or the generic (\S\ref{sec:pacc_intuitions}). We note that previous work found that
the surprisal of the whole sequence is not sensitive to the effect of quantifiers \citep{collacciani2024quantifying} and we replicate this finding (see Appendix \ref{apx:ppl}). 

\section{Experiments}
\label{sec:experiments}

The experiments that follow use p-acceptability (defined in \S\ref{sec:acceptability}) to study quantification and generics through language models. 
First, we validate
that p-acceptability describes quantification in \textsc{ConGen} and \textsc{GenericsKB} (\S\ref{sec:pacc_intuitions}). Then, we explore three aspects of generics: their implicit quantificational strength (\S\ref{sec:implicit_quantification}), their context-sensitivity (\S\ref{sec:context_sensitivity}), and their role in stereotypes (\S\ref{sec:striking}).


We use three state-of-the-art open-source language models of increasing size: \textsc{Mistral-7B}, \textsc{Mistral-8\texttimes7B} and \textsc{Mistral-8\texttimes22B} \citep{jiang2023mistral}. Additional details on the models used are available in Appendix \ref{apx:language_models}.

Because our focus is on bare plural generics, we filter out of \textsc{GenericsKB} those generics that are not bare plural
 and call this subset \textsc{GenericsKB-BP} ($N=570358$).  Implementation details can be found in Appendix \ref{apx:experimental_setup}.

\subsection{Can p-acceptability describe quantification?}
\label{sec:pacc_intuitions}

Quantifiers specify a prevalence relation between members of a kind and a property. In terms of this relation, we would expect \textit{all} and \textit{most} to be interchangeable in many contexts, likewise for \textit{most} and \textit{some} but never for \textit{all} and \textit{some}.
This experiment recovers these commonsense intuitions of quantification with p-acceptability. 

\paragraph{Experimental setup.} For each sentence in \textsc{ConGen} and \textsc{GenericsKB-BP}, we build the set of variations and get the p-acceptable quantifier. We plot these p-acceptability percentages against the original quantifiers of the sentences, that is, how often each quantifier 
makes the property tokens easiest to predict for the language model.

\begin{figure}
  \centering
  \includegraphics[width=\linewidth]{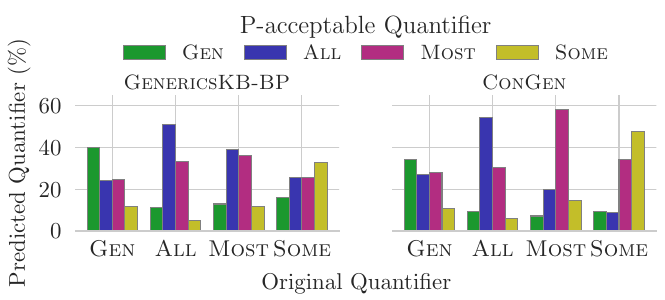}
  \caption{P-acceptable quantifiers on both datasets correspond to semantic intuitions (\textsc{Mistral-7B})}
  \label{fig:gkb_congen_pacc}
\end{figure}

\paragraph{Results.} In both datasets, the most prevalent p-acceptable quantifier corresponds to the original quantifier (Figure \ref{fig:gkb_congen_pacc}). 
In \textsc{GenericsKB} data, the distinctions are less clear, with bigger confusion between, for example, \textit{all} and \textit{most}.

For sentences that were originally generic, \textit{all} and \textit{most} are the most prevalent wrongly p-accepted quantifiers. This agrees with most generics being majority generics.

The prevalence of \textit{all} in originally \textit{some} sentences from \textsc{GenericsKB-BP} seems counter-intuitive. We believe that this is due to noise in the dataset\footnote{For example, the three first \textit{some} sentences in \textsc{GenericsKB} are \textit{Some aardvarks detect predators}, \textit{Some aardvarks dig holes} and \textit{Some aardvarks dig own burrows}. These feel more appropriate for \textit{most} or \textit{all} quantification.}, rather than the metric. On \textsc{ConGen}, p-acceptability recovers an intuitive profile for \textit{some} sentences: \textit{some} is the most prevalent quantifier and is mostly confused with \textit{most}, rarely with \textit{all} or the generic.

P-acceptability captures semantic intuitions on quantification across both datasets. In what follows, we use p-acceptability to investigate some aspects of how speakers \textit{use} generics.


\subsection{What is the implicit quantification of generics?}
\label{sec:implicit_quantification}

Although generic sentences present no overt quantification operator, we can investigate which quantifier better describes the kind-property relationship expressed in a generic with p-acceptability. Given a generic sentence, we study its \textit{implicit quantification} by finding the p-acceptable explicit quantifier. 

\paragraph{Experimental setup.} In this experiment we consider generics from \textsc{ConGen} and \textsc{GenericsKB-BP}. We compute the p-acceptability excluding \textsc{Gen} from the candidate quantifiers and only considering \textit{all}, \textit{most} and \textit{some} as possible options for quantification.

\begin{figure}
  \centering
  \includegraphics[width=.75\linewidth]{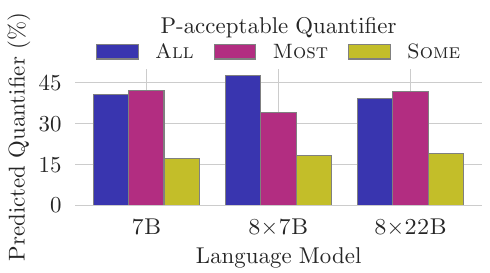}

  \caption{Implicit quantification in \textsc{ConGen} generics across \textsc{Mistral} models.}
  \label{fig:congen_all_no_gen}
\end{figure}

\paragraph{Results.} Across all three models (Figure \ref{fig:congen_all_no_gen}), \textit{all} and \textit{most} are the most p-acceptable implicit quantifiers at a prevalence of 40\% each. Around 18\% of sentences are consistently quantified as \textit{some}. 

For the \textsc{7B} and \textsc{8x22B} models, \textit{most} is the most prevalent quantifier, which mirrors the fact that generic sentences often express properties shared by a majority of members of a kind. Nevertheless, \textit{all} has comparable or even bigger prevalence for \textsc{Mistral-8\texttimes7B}.
Although we expect \textit{all} to be the implicit quantifier for quasi-definitional generics, the observed high prevalence
of \textit{all} 
suggests that language models also model the generic overgeneralisation effect (\S\ref{sec:generics_and_human_bias}), as found in other studies \citep{allaway2024exemplars_llms}.
A more fine-grained annotation is needed to verify this on \textsc{ConGen} data.

\begin{table}[!t]
  \small
  \centering
  \begin{tabular}{p{0.9\linewidth}}
    \multicolumn{1}{c}{\textsc{All}} \\
      \midrule 
      Aquatic crustaceans have gills for breathing. \\
      Dead plants contain vital substances beyond just carbon. \\
      Omnivores eat both plants and animals. \\
      \\
      \multicolumn{1}{c}{\textsc{Most}} \\ 
      \midrule 
      Narcotics cause a good deal of vasodilation. \\
      Banana plants have a lot of root exudates. \\
      People adapt to total blindness. \\
      \\
      \multicolumn{1}{c}{\textsc{Some}} \\ 
      \midrule 
      Ocean currents carry water over long distances. \\
      Berries are toxic to humans but loved by birds. \\
      Plastics bind heavy metals. \\
  \end{tabular}
  \caption{Examples of naturally occurring generics from \textsc{ConGen} with different implicit quantifications. (p-acceptability from \textsc{Mistral-7B}).}
  \label{tab:examples_implicit}
\end{table}

We take those generics with \textit{some} as the p-acceptable quantifier to be weak generics.
We observe close to 20\% of weak generics across all models both in \textsc{ConGen} (Figure \ref{fig:congen_all_no_gen}) and \textsc{GenericsKB-BP} (Figure \ref{fig:gkb_all_no_gen}). To the best of our knowledge, this is the first estimation of the prevalence of weak generics in natural language.
Table \ref{tab:examples_implicit} shows examples of generics with different implicit quantifications (also in Appendix \ref{apx:data_samples}, Table \ref{tab:weak_generics_examples}). 

\subsection{Are generics context-sensitive?}
\label{sec:context_sensitivity}

Semantic theories in philosophy of language hypothesize that the context of generic sentences determines the semantic content of \textsc{gen} (\S\ref{sec:semantic_theories}). We 
quantify the effect of different context windows on implicit quantification using p-acceptablility and
the multi-sentence contexts in \textsc{ConGen}.

\paragraph{Experimental setup.} For each sentence in \textsc{ConGen}, we compute the p-acceptable quantifier at increasing sizes of left-side context. We increase the context size in chunks of 4 tokens, irrespective of word or sentence boundaries (Table \ref{tab:cricket_context}).

We measure the percentage of correct predictions by p-acceptability as instances where it recovers the original quantifier. For sentences that are originally generics, we also replicate 
this
setup (excluding \textsc{gen} from the candidate quantifiers) for different left-context windows.

\begin{figure}
  \centering
  \includegraphics[width=.9\linewidth]{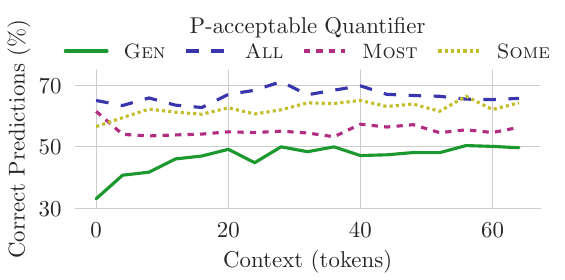}
  \caption{Percentage of correct p-acceptable quantifiers with different contexts. (\textsc{Mistral-8\texttimes22B})}
  \label{fig:context_sensitivity}
\end{figure}

\paragraph{Results.} Figure \ref{fig:context_sensitivity} shows the percentage of correct predictions for each original quantifier (e.g. green corresponds to the percentage of times \textit{gen} is p-accepted on generic sentences at each context length). In originally generic sentences, we have a 20\% increase in accuracy across the first 20 tokens of context, which roughly correspond to the preceding sentence.
For explicitly quantified expressions, context does not improve the accuracy of p-acceptability as much as for generics.

As control, we replicate the experiment with random context sampled from other documents with the same original source (\textsc{Dolma} or Reddit) and find no improvement on any quantifier, including \textsc{gen}. Details are available in Appendix \ref{apx:random_contexts}.

\begin{figure}
  \centering
  \includegraphics[width=.9\linewidth]{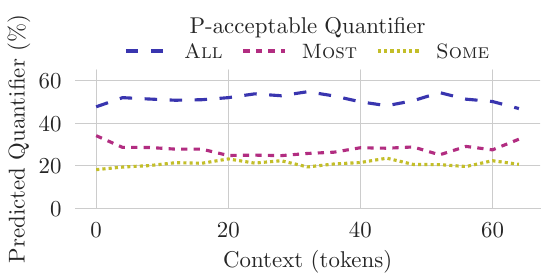}
  \caption{Implicit quantification with different left-side contexts on generic sentences from \textsc{ConGen}. (\textsc{Mistral-8\texttimes22B})}
  \label{fig:gen_ctx_no_gen}
\end{figure}

We investigate if the increase in accuracy that context has on generic sentences is related to their implicit quantification. In Figure \ref{fig:gen_ctx_no_gen} the relative percentages of each quantifier are mostly unaffected by context, with a slight increase of \textit{all} and decrease of \textit{most}.

For those samples where context is needed for p-acceptability to predict the correct quantifier, we define the \textit{minimal context} as the smallest context needed for the correct prediction. We find a very low presence of quantifiers in the minimal contexts of generic sentences.
A preliminary analysis of the linguistic characteristics of these contexts is available in Appendix \ref{apx:minimal_contexts}. 


\subsection{Are stereotyping generics different?}
\label{sec:striking}

\begin{figure*}
  \centering
  \includegraphics[width=.9\linewidth]{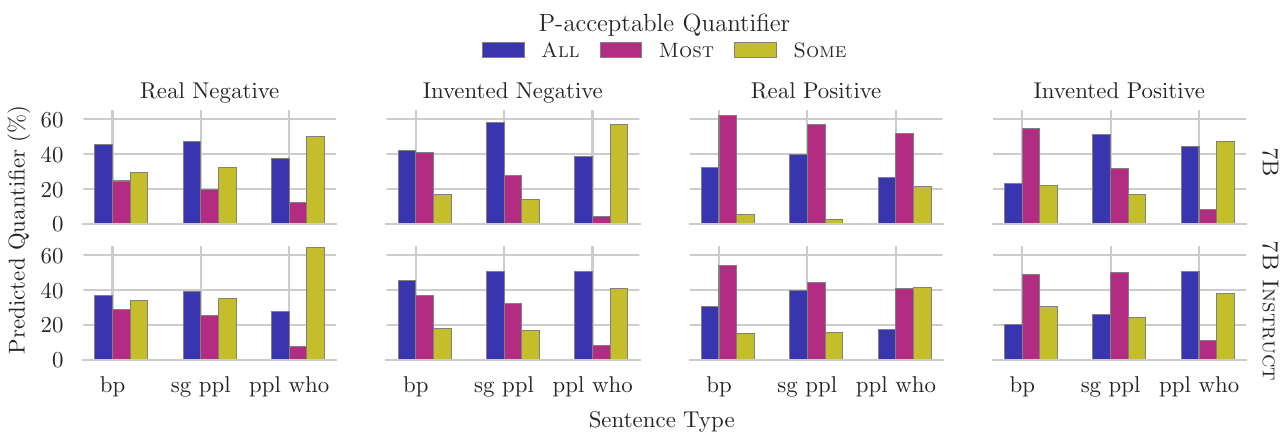}
  \caption{Different p-acceptability rates for each paraphrase of stereotyping generic sentences for \textsc{Mistral-7B} and \textsc{Mistral-7B Instruct}. Paraphrases are indicated as \textit{bp} (bare plural), \textit{sg ppl} (singular $+$ `people') and \textit{ppl who} (`People who are' $+$ singular).}
  \label{fig:striking_histplot}
\end{figure*}

Stereotypes are often expressed linguistically through striking generics where the subject is a social group. This is partly because, when dangerous properties are predicated, humans perceive them as more prevalent than they really are~\citep{cimpian2010generic,leslie2017originalsin}. In the following experiment, we study the implicit quantification in language models of negative and positive stereotypes. 

\paragraph{Experimental setup.} To study the implicit quantification in this subset of generics, we collect a small dataset of stereotypes ($N=504$) divided into \textit{real} (the subject-property is a real-world stereotype) and \textit{invented} sentences (the subject is an invented word that morphologically resembles a demonym).

We extract real negative stereotypes from the Social Bias Frames dataset \citep{sap2020social}, a collection of offensive texts annotated with implied stereotypes. For real positive stereotypes, we generate samples based on tradition and culture for different social groups. The invented sentences are built by combining invented demonyms with a list of negative and positive predicates (e.g., \textit{craguils are murderers} or \textit{corriards are warm and
hospitable}). Details are available in Appendix \ref{apx:stereotypes} and Table \ref{tab:stereotypes_data}.

To further explore how effective purely linguistic strategies are at mitigating the bias in striking generics~\citep{leslie2017originalsin,carnaghi2008nomina,gelman1999carrots}, we generate the following three paraphrases for each social group in a stereotype:
bare plural (\textit{catalans are lovely}), singular $+$ `people' (\textit{catalan people are lovely}) and
`people who are' $+$ singular (\textit{people who are catalan are lovely}).

We compare the results between the pre-trained and instruction tuned versions of \textsc{Mistral-7B}, as one objective of language model designers when instruction-tuning models is to mitigate social biases \citep{zhang2024instructiontuninglargelanguage}.

\paragraph{Results.} Figure \ref{fig:striking_histplot} reports the percentage of p-acceptable quantifiers for each paraphrase and type of stereotyping generic. 
For negative stereotypes, \textit{all} is the predominant quantifier. This aligns with
the theoretical and empirical observation that speakers use universal quantification with this subset of striking generics~\citep{cimpian2010generic}.
Note that with
the \textit{people who are} paraphrase
this is not the case;
we observe a stark contrast, where \textit{some} is the most prevalent p-acceptable quantifier. 
The instruction-tuned model predicts more \textit{some} and less \textit{all}.
Interestingly, for invented negative cases even in the \textit{ppl who} paraphrase, \textit{all} is the most prevalent quantifier. 

In contrast to the negative stereotypes, the
predominant quantifier is \textit{most} for positive stereotypes.
This further supports hypotheses that the implicit universal quantification of negative stereotypes is due to the strikingness of the predicate.




\section{Discussion}
\label{sec:discussion}

In this work, we study different aspects of generics and quantified sentences through language models.
We now discuss our results in relation to existing theories of generics.

\paragraph{Weak generics.} Weak generics are central to discussions of generics in philosophy of language. On the one hand, \citet{generics_08_leslie} uses the prevalence of striking generics to support the idea that generics express primitive psychological generalisations. In contrast, \citet{sterken2015errortheory} proposes an error theory where striking generics are false. Additionally, \citet{gustafsson2023truth} examines weak (striking) generics and argues that generics are more heterogeneous than what the previous theories take them to be.
Although these works use striking generics as their running examples, they pay little attention to \textit{how many} generics are actually weak nor 
what the non-striking weak generics \textit{look like}.

In our experiments on implicit quantification (\S\ref{sec:implicit_quantification}) we characterize weak generics as those generics implicitly quantified with \textit{some}.
If we take the $437414$ generics in \textsc{ConGen} and \textsc{GenericsKB-BP} combined to be a representative sample of language, between 18\% and 23\%\footnote{These are the percentages of generic sentences with the p-acceptable quantifier 
\textit{some} with \textsc{Mistral-8\texttimes22B} on \textsc{ConGen} and \textsc{GenericsKB-BP} respectively.} of generics are weak generics. Manual examination of these weak generics reveals that they are mostly non-striking (Tables \ref{tab:examples_implicit} and \ref{tab:weak_generics_examples}). 

These results suggest that \textit{non-striking weak generics are common in language use}: almost 1 in 5 generics we analyze expresses a weak generalisation.
Our notion of implicit quantification not only can give an estimation of \textit{how many} weak generics are there, but, when applied to \textsc{ConGen}, yields explicit examples of weak generics in-context. This is a new resource for theorists to draw from.





\paragraph{Context-sensitivity.}
The degree of context-sensitivity of the semantic content of \textsc{gen} is a controversial topic in the literature. 
The \textit{generics-as-defaults} view posits 
stable, non-context-sensitive content
whereas the \textit{contextualist} view claims a distinct and strong context-sensitivity. Recently, \citet{gencontext_23_almotahari} argued that both views are compatible, by attributing some context-sensitivity to psychologically salient features in \textit{generics-as-default}.

In our experiments (\S \ref{sec:context_sensitivity}), increasing context improves the accuracy of p-acceptability for generic sentences much more than for explicitly quantified sentences. Additionally, when considering the implicit quantification of generic sentences, the prevalence of each quantifier is not affected by the context. These results suggest that \textit{generics are context-sensitive in a way that determiner quantifiers are not}.

In this work, we limit our contribution to showing that some dynamics of context-sensitivity in generics can be revealed with language models, and leave further inquiry, both empirical and theoretical, for future work.

\paragraph{Stereotypes.} The use of generics to express stereotypes is a much explored topic both in philosophy of language and experimental psychology~\citep{cimpian2010generic,leslie2017originalsin}. The following two hypotheses are central to the discussion: (\emph{i}) humans interpret negative stereotypes as universally quantified and (\emph{ii}) how speakers express a stereotype changes its the perceived quantificational force.

In the context of stereotypes, we take implicit quantification as a measure of bias: interpreting stereotypes as universally quantified is a sign of essentialization and prejudice. An unbiased system should quantify stereotypes as \textit{some}, sometimes \textit{most}, but never \textit{all}.

Our results (\S\ref{sec:striking}) are congruent with both hypotheses. We find that negative stereotypes are overwhelmingly implicitly quantified as universals (\textit{all}), whereas for positive stereotypes the most p-accepted quantifier is \textit{most}. For those same negative stereotypes, when paraphrased as \textit{ppl who}, \textit{some} becomes increasingly prevalent, suggesting an existential perception of quantification.

The study of social bias in language models is important in order to make them fair and safe to use. Bias mitigation is often a priority of language model designers when instruction tuning. For real stereotypes, the instruction-tuned model predicts less \textit{all} and more \textit{some} than the base model, which suggests a success in bias mitigation. 
Nevertheless, on invented negative stereotypes, the instruction model predicts a large amount of \textit{all}, even for the \textit{ppl who} case.
Although our experiments are not on generation, this raises doubts on the effectiveness of instruction tuning for bias mitigation, especially when new social kinds are combined with striking properties.

\section{Conclusion}
\label{sec:conclusion}

Generics are similar to quantifiers, yet speakers use them in logically inconsistent ways. In this work, we study the dynamics of quantification on generic and quantified sentences through language models.

To do so, we introduce a new dataset (\textsc{ConGen}) and metric (p-acceptability). With these tools we estimate the prevalence of weak generics, identify a distinct context-sensitivity in generics and show how linguistic strategies can help mitigate stereotypes.

We believe our findings and methodology open new doors for research on generics and quantification in language.



\section{Limitations}

\paragraph{Language models.} Even though we use SOTA open-source language models for our experiments, currently all competitive language models are trained for profit rather than research; this inevitably hinders any research effort. 

Compute limitations mean we run \textsc{Mistral-8\texttimes7B} 8-quantized and \textsc{Mistral-8\texttimes22B} 4-quantized. There is empirical evidence that quantization does not have a big impact on performance for \textsc{Mistral} models across a wide range of tasks \citep{badshah2024quantizedmistral}.

We test on a family of models with the autoregressive transformer architecture. Exploration of how other autoregressive families or architectures (such as MAMBA or diffusion) model quantification and generics is left for future work.

\paragraph{Metric.} The p-acceptability metric, as defined, is specific to English sentences with a simple structure, as the bare plural does not exist in many other languages.

\paragraph{Implicit quantification} The implicit quantification in generic sentences could be more closely related to adverbial quantifiers than to the determiner quantifiers we study \citep{kirkpatrick2024quantificational}. Future work will need to expand both the metric and dataset to include other quantifiers like \textit{many}, \textit{every} or \textit{few}, in order to get a more comprehensive picture of quantification and its relation to generics.

\paragraph{Data.} Sentences in \textsc{ConGen} are first collected from \textsc{Dolma} and Reddit by a classifier trained to identify generic and quantified sentences. Performance and biases of this classifier are not well explored in this work and could affect the significance of the sampling. Another source of bias in \textsc{ConGen} is that the first author annotates most of the data. We plan on adressing these issues and expanding \textsc{ConGen} in future work.

In the stereotypes collection we build, the stereotypes are derived from a dataset based on American Twitter, which means they are centered around American and Western culture and prejudice.

\paragraph{Context.} The concept of context can broadly mean three things in the philosophy of language literature: the spatial and temporal context of the utterance, the subjective context of the speaker (such as intentions) or the linguistic context (previous utterances). In this work, we assume linguistic context as the only source of context-sensitivity, as we study how language models model the context. If generics were context-sensitive in ways that are not expressed or conveyed in language, our methodology could not capture it.

\section*{Acknowledgements}
This work was funded by UK Research and Innovation (UKRI) under the UK government’s Horizon Europe funding guarantee [grant number 10039436].

The first author would like to thank Mahrad Almotahari for introducing him to the topic of generics and Dan Lassiter, Annie Bosse,  Nicolas Navarre, Tristan Baujault-Borresen, Guillem Ramírez and Aina Centelles for their proof-reading, corrections and discussions. 

We also thank the anonymous reviewers for their comments on the manuscript.

\nocite{stanza}
\nocite{SciencePlots}
\nocite{baumgartner2020pushshiftreddit}
\nocite{labelstudio}
\nocite{brandom1994explicit}
\nocite{wittgenstein_1953_philosophical}
\nocite{heidegger1927being}

\bibliography{custom}

\clearpage

\appendix

\renewcommand{\thefigure}{\thesection.\arabic{figure}}
\renewcommand{\thetable}{\thesection.\arabic{table}}
\setcounter{figure}{0}  
\setcounter{table}{0}  

\section{Language Models}
\label{apx:language_models}

In our experiments we use autoregressive transformers from the \textsc{Mistral} family \citep{jiang2023mistral}. Table \ref{tab:lms} compares their size and performance in the MMLU benchmark \citep{hendrycks2020mmlu}, which test the performance of language models across a wide range of topics and tasks.

\begin{table}[h]
  \centering
  \small
  \begin{tabular}{lll}
    \multicolumn{1}{l}{Model name} & \multicolumn{1}{l}{Active params.} & \multicolumn{1}{l}{MMLU ($\uparrow$)}\\ \midrule
    {\sc Mistral-7B} & 7 B & 62.5\%  \\
    {\sc Mixtral-8\texttimes7B}  & 12.9 B & 70.6\% \\
    {\sc Mixtral-8\texttimes22B} & 39 B & 77.8\% \\
    {\sc GPT-3.5 turbo} & {\small undisclosed} & 70.0\% \\
  \end{tabular}
  \caption{Comparison of size and performance  of the  language models used, with \textsc{gpt-3.5 turbo} (ChatGPT) for comparison.}
  \label{tab:lms}
\end{table}

\section{\textsc{ConGen} dataset}
\label{apx:congen}

\paragraph{Data sources.} We use a sample of 1.45 million documents (over 100 million sentences) from \textsc{Dolma} (\url{https://huggingface.co/datasets/andersonbcdefg/dolma_sample}). We scrape the comments written in June and July 2024 from Reddit using the \url{pullpush.io} API \citep{baumgartner2020pushshiftreddit} from a list of popular subreddits. The context for the Reddit comments is the comment itself, not a thread with other responses.

\paragraph{Generics classifier training.} For the \textsc{Gen-a-Tonic} dataset, we label as negative examples sentences with a i2d2 score $<0.6$ or that do not match the following regex pattern: \lstinline[basicstyle=\ttfamily\small]{|can | may | should | you | before|}. We also remove adverbial quantifiers like \texttt{typically} and \texttt{generally}. For \textsc{GenericsKB} we use the scores of the original dataset as labels. The resulting training dataset has $N=9414001$ samples.

We train a \textsc{RoBERTa}-base model \citep{liu2019roberta} with a sequence classification head on this dataset for one epoch with learning rate $2\times 10^{-5}$.

\paragraph{Candidate sentences and annotation.} To reduce candidate sentences from millions to few thousands, we filter the sentences with the \textsc{RoBERTa} classifier (score $>0.7$) and the regex pattern below, which includes words that are often incompatible with the generic and quantified bare plurals we study:
\begin{lstlisting}[breaklines=true, basicstyle=\ttfamily\small]
is | may | can | should | would | must | have to | will | you |^i | were | was | many | we | they | ought | your |^[^ ]+ of | us | \? | this | that | those | these | all in all |,|^the |^a |than 
\end{lstlisting}
We also filter out sentences in the passive voice \citep{sepehri2023passive}.

The candidate sentences are annotated by the authors using Label Studio \citep{labelstudio}.

\section{P-acceptabilty with $H$ and $H_p$}
\label{apx:ppl}

We compare the notion of acceptability in equation \ref{eq:p-acceptability} using the surprisal over the whole sequence ($H$) or the surprisal on the property tokens ($H_p$). We observe that the surprisal over the whole sequence is not able to correctly predict generic sentences unless we provide big context windows.

\begin{table}[h!]
  \centering
  \small
  \begin{tabular}{ccc}
   Context (tokens) & $H$ & $H_p$ \\
    \midrule
    0 & 0.007 & 0.34 \\
    32 & 0.03 & 0.43 \\
    128 & 0.32 & 0.45 \\


  \end{tabular}
  \caption{Accuracy on originally generic sentences of p-acceptability using the entropy over the whole sequence ($H$) or on the property tokens ($H_p$); $H$ needs a big context window to be sensitive to generics (\textsc{Mistral-7B}).}
  \label{tab:ppl_vs_placc}
\end{table}

\section{Experimental setup}
\label{apx:experimental_setup}

To use \textsc{GenericsKB} in our setting, we select only bare and quantified plurals. The quantified samples are already labeled in the original dataset (\textit{all}, \textit{most} and \textit{some}). We select the bare plurals from the generic samples by finding sentences with plural verbs in the present indicative with Stanza \citep{stanza}. We call this filtered version \textsc{GenericsKB-BP}.

\section{Implicit quantification}

\begin{figure}[h!]
  \centering
  \includegraphics[width=.8\linewidth]{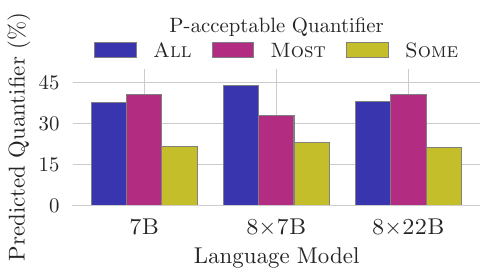}

  \caption{Implicit quantification landscape in \textsc{GenericsKB-BP} generics across \textsc{Mistral} models.}
  \label{fig:gkb_all_no_gen}
\end{figure}

In \textsc{GenericsKB-BP}, the percentage of sentences implicitely quantified as \textit{some} is slightly higher, being 23\% for \textsc{Mistral-8\texttimes22B} (Figure \ref{fig:gkb_all_no_gen}).

\section{Context-sensitivity with random contexts}
\label{apx:random_contexts}

\begin{figure}[h!]
  \centering
  \includegraphics[width=.9\linewidth]{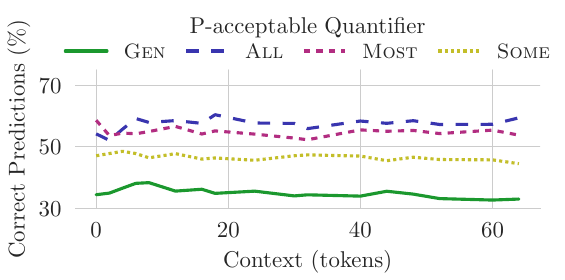}
  \caption{Context-sensitivity results with a random context. (\textsc{Mistral-7B}).}
  \label{fig:random_context}
\end{figure}

We reproduce the experiment on context-sensitivity by using random contexts. For each sentence, we sample a context from the same source (\textsc{Dolma} or Reddit) and find the p-aceptable quantifier. Random contexts do not improve the accuracy of p-acceptability (Figure \ref{fig:random_context}).

\section{Minimal contexts}
\label{apx:minimal_contexts}

We search minimal contexts of each original quantifier for the following linguistic characteristics: \textit{quantifiers} (from a list of 41 in total), \textit{noun last} (whether the last word in the context is a noun \citep{stanza}), \textit{question} (context contains a question), \textit{all}, \textit{most} and \textit{some} (context contains the respective quantifier). In Table \ref{tab:minimal_context_analysis}, we compare how these characteristics change from all contexts in \textsc{ConGen} (regardless of the p-accepted quantifier) to minimal contexts.

There are $N=2314$ original contexts (as some samples in \textsc{ConGen} have no left context), which are truncated to the first 64 tokens. For minimal contexts the sample size is $N=512$, which are the samples that have an incorrect p-accepted quantifier without context but a correct one after some tokens are added (the minimal context).

The 41 quantifiers we match in the contexts are: \texttt{\small all, some, each, every, no, much, more, most, less, few, several, many, enough, little, various, always, usually, often, frequently, sometimes, occasionally, seldom, rarely, never, almost, nearly, hardly, scarcely, barely, completely, entirely, totally, absolutely, partly, largely, mostly, entirely, extremely, exceptionally, especially, particularly}.

\begin{table}[h!]
  \small \centering
  
  \begin{tabular}{lcccc} 
    & \textsc{Gen} & \textsc{All} & \textsc{Most} & \textsc{Some} \\ 
    \midrule 
    \vspace{1pt} 
    \textit{quantifier} & $52|11$ & $59|17$ & $58|24$ & $61|23$ \\ 
    \vspace{1pt} 
    \textit{noun last} & $63|70$ & $59|54$ & $60|60$ & $65|53$ \\
    \vspace{1pt} 
    \textit{question} & $20|37$ & $35|18$ & $27|29$ & $17|16$ \\
    \vspace{1pt} 
    \textit{all} & $9|2$ & $13|7$ & $8|3$ & $9|4$ \\ 
    \vspace{1pt} 
    \textit{most} & $5|1$ & $7|2$ & $11|5$ & $11|3$ \\ 
    \vspace{1pt} 
    \textit{some} & $9|0$ & $8|0$ & $10|2$ & $13|4$ \\
   
  \end{tabular} 
  \caption{Percentage (\%) of contexts that present each linguistic characteristic for the full context (left) and minimal context (right). (\textsc{Mistral-8\texttimes22B})}
  \label{tab:minimal_context_analysis} 
\end{table}

\section{Stereotypes}
\label{apx:stereotypes}

\paragraph{Real negative stereotypes.} We manually select stereotypes from the \texttt{targetStereotype} column in the Social Bias Frames dataset \citep{sap2020social} and parse them into singular, plural and predicate (to be able to automatically construct the paraphrases).

\paragraph{Real positive stereotypes.} We use ChatGPT to prompt different backend models (\textsc{Gpt4} and \textsc{Gpt3.5}) for predicates that refer to tradition and food from different countries, races and religions. The data is then manually revised and edited.

\paragraph{Invented data.} The invented data is built by combining a list of invented words morphologically resembling demonyms with two lists of predicates, one for negative and the other for positive stereotypes.

Invented demonyms: \texttt{\small slubrack, craguil, flirel, corrish, xuni, rorian, kryghin, dunfamese, zorvish, lampagurd}. 

Positive predicates: \texttt{\small are generous, help out their colleagues, love their family, work really hard, cook good food, are warm and hospitable, care for their community, are smart, help people in need, are creative, care for their friends and family, are great musicians}.

Negative predicates: \texttt{\small are aggressive, smell bad, are alcoholics, are terrorists, are murderers, kill other people, are perverts who molest children, are sexual predators, carry infectious diseases, are violent drunkards, are arrogant and narcissistic, are greedy and stingy with money}. 

\paragraph{On synthetic data.} The sentences of this stereotype collection are synthetic, as we create the paraphrases by parsing the subject and predicate of stereotypes. We assume they are generics because we use generics to express stereotypes \citep{rhodes2012cultural}, but as they are not naturally occurring, we do not know if the stereotype might be expressed by explicit quantification by native speakers. The implicit quantification results should be interpreted as quantification of the stereotype, rather than of the generic sentence.

\begin{table}[h!]
  \small
  \centering
  \begin{tabular}{lcc}
    & Negative & Positive  \\
    \midrule
    Real & 144 & 120 \\
    Invented & 120 & 120 \\
  \end{tabular}
  \caption{Statistics of the stereotypes data.}
  \label{tab:striking_data}
\end{table}

\section{Data samples}
\label{apx:data_samples}

\begin{table*}
  \centering
  \begin{tabular}{p{4.8cm} p{4.8cm} p{4.8cm}}
  \multicolumn{1}{c}{\textsc{All}} & \multicolumn{1}{c}{\textsc{Most}} & \multicolumn{1}{c}{\textsc{Some}} \\
  \toprule
  Commercial dish detergents are chemical sanitizers. & Roots do not grow through concrete. & Bee fly larvae are parasitic and eat the larvae of other insects. \\
  MCTs are fatty acids. & Parents serve healthy food. & Mother velvet worms carry their babies for up to 15 months. \\
  Zygotes form after fertilization. & Tomatoes are full of glutamic acid. & Americans use real guns. \\
  Healthy gums need collagen. & Bears only attack when starving or threatened. & Birds grind seeds with pebbles in their gizzard. \\
  Plants don’t need a central nervous system to feel pain. & Swimmers have long torsos. & Plants need mushrooms to grow. \\
  Humans have cognitive bias. & Dinosaurs do not eat humans. & Cellular components can self-organize into higher order structures. \\
  Mycotoxins are poisonous byproducts of fungi. & Rays lay eggs internally for about a year. & Old people are savage. \\
  Magnesium-rich almonds transport calcium. & Hyraxes have stumpy toes with hoof-like nails. & Crocodiles also attack and eat sharks. \\
  Mitochondria are not cells & Leaves grow on trees. & MAGA Maggots thrive on decay. \\
  Dental caries are tooth decay. & Beavers are herbivorous semi-aquatic. & Tomato plants grow anti-predation spikes. \\
  Weimaraners are canine athletes. & Omega-6’s are inflammatory acids. & Grains damage germ plasm. \\
  Succulents release CO2 in the day and O2 at night. & Cows are large animals. & Estate sales are a gold mine. \\
  Sundews have mucilage glands. & Alveoli manage oxygen and CO2 in blood. & Stimulants work by increasing dopamine. \\
  Black men are men of culture & Bed bugs have distinct flat brown bodies. & Caterpillars naturally disperse by wind. \\
  Antiviral drugs stop HSV replication. & Red blood cells transport oxygen. & Modern day birds descend from dinosaurs. \\
  Deciduous plants lose leaves in fall. & House mice don't survive winter outdoors. & Dental sealants work as protection. \\
  B cells produce antibodies. & Safety regulations are written in blood & Genes act differently depending on environment. \\
  Unicellular organisms have just one cell. & Window world are crooks & Human bodies are liquid. \\
  Cats use their noses for communication. & Marine turtles have flippers, spend life in ocean & Foods cause acne breakouts. \\
  \bottomrule

  \end{tabular}
  \caption{Naturally occuring generics that are classified implicitely quantifying as \textit{all}, \textit{most} or \textit{some} with p-acceptability. Samples from \textsc{ConGen}. (\textsc{Mistral-7B})}
  \label{tab:weak_generics_examples}
\end{table*}

\begin{table*}[h!]
  \centering
  \begin{tabular}{p{1cm} p{10cm}}
    \multicolumn{1}{l}{\bf Context Tokens} & \multicolumn{1}{l}{\bf Context} \\
    \midrule
    4  & the spider. \\
    8  & this surprising reaction of the spider. \\
    12 & me an explanation for this surprising reaction of the spider. \\
    16 & not long ago gave me an explanation for this surprising reaction of the spider. \\
    20 & of Natural History, not long ago gave me an explanation for this surprising reaction of the spider. \\
    24 & at the American Museum of Natural History, not long ago gave me an explanation for this surprising reaction of the spider. \\
    28 & , spider authority at the American Museum of Natural History, not long ago gave me an explanation for this surprising reaction of the spider. \\
    32 & . Gertsch, spider authority at the American Museum of Natural History, not long ago gave me an explanation for this surprising reaction of the spider. \\
    36 & . Willis J. Gertsch, spider authority at the American Museum of Natural History, not long ago gave me an explanation for this surprising reaction of the spider. \\
    40 & rider. Dr. Willis J. Gertsch, spider authority at the American Museum of Natural History, not long ago gave me an explanation for this surprising reaction of the spider. \\
    44 & whip toward the rider. Dr. Willis J. Gertsch, spider authority at the American Museum of Natural History, not long ago gave me an explanation for this surprising reaction of the spider. \\
    48 & and ran along the whip toward the rider. Dr. Willis J. Gertsch, spider authority at the American Museum of Natural History, not long ago gave me an explanation for this surprising reaction of the spider. \\
    52 & leaped upon it and ran along the whip toward the rider. Dr. Willis J. Gertsch, spider authority at the American Museum of Natural History, not long ago gave me an explanation for this surprising reaction of the spider. \\
    56 & the wolf spider leaped upon it and ran along the whip toward the rider. Dr. Willis J. Gertsch, spider authority at the American Museum of Natural History, not long ago gave me an explanation for this surprising reaction of the spider. \\
    60 & inches from it, the wolf spider leaped upon it and ran along the whip toward the rider. Dr. Willis J. Gertsch, spider authority at the American Museum of Natural History, not long ago gave me an explanation for this surprising reaction of the spider. \\
    64 & the earth a few inches from it, the wolf spider leaped upon it and ran along the whip toward the rider. Dr. Willis J. Gertsch, spider authority at the American Museum of Natural History, not long ago gave me an explanation for this surprising reaction of the spider. \\
    \bottomrule
    \end{tabular}

  \caption{Example of contexts from 4 to 64 tokens for the \textsc{ConGen} sample \textit{All wolf spiders are sensitive to vibrations in the ground.}}
  \label{tab:cricket_context}
\end{table*}

\begin{table*}
  \small
  \centering
  \begin{tabular}{p{2.5cm} p{3cm} p{3cm} p{3cm}}
    \multicolumn{1}{l}{} & \multicolumn{1}{c}{\bf Bare plural} & \multicolumn{1}{c}{\bf Singular + people} & \multicolumn{1}{c}{\bf People who are + singular} \\
  \midrule
  \multirow{2}{*}{\bf Real Negative} & liberals are corrupt & liberal people are corrupt & people who are liberal are corrupt \\ 
  & democrats are cheaters & democrat people are cheaters & people who are democrats are cheaters \\ \midrule
  \multirow{2}{*}{\bf Invented Negative} & slubracks smell bad & slubrack people smell bad & people who are slubrack smell bad \\
  & craguils are murderers & craguil people are murderers & people who are craguil are murderers \\ \midrule
  \multirow{2}{*}{\bf Real Positive} & spaniards love gazpacho in the summer & spanish people love gazpacho in the summer & people who are spanish love gazpacho in the summer \\
  & germans celebrate with lederhosen and dirndls & german people celebrate with lederhosen and dirndls & people who are german celebrate with lederhosen and dirndls \\ \midrule
  \multirow{2}{*}{\bf Invented Positive} & flirels are smart & flirel people are smart & people who are flirel are smart \\
  & corriards are warm and hospitable & corriard people are warm and hospitable & people who are corriard are warm and hospitable \\ \midrule
  \end{tabular}
  \caption{Samples from the stereotypes dataset.}
  \label{tab:stereotypes_data}
\end{table*}

\end{document}